\documentclass[10pt,twocolumn,letterpaper]{article}

\usepackage{cvpr}              

\usepackage{amssymb}
\usepackage{amsmath}

\usepackage{amsmath,amsfonts}
\usepackage{algorithmic}
\usepackage{algorithm}
\usepackage{array}

\usepackage{textcomp}
\usepackage{stfloats}
\usepackage{url}
\usepackage{verbatim}
\usepackage{graphicx}

\usepackage{booktabs}
\usepackage{multicol}
\usepackage{multirow}

\usepackage{colortbl}  
\usepackage{xcolor}
\usepackage{color}
\usepackage{xspace}
\usepackage{pifont}
\usepackage{bbding}

\usepackage{enumitem}

\usepackage{tabularx}

\usepackage[bold]{hhtensor}

\usepackage{wrapfig}
\definecolor{cvprblue}{rgb}{0.21,0.49,0.74}
\usepackage[pagebackref,breaklinks,colorlinks,allcolors=cvprblue]{hyperref}
\usepackage[capitalize]{cleveref}
\crefname{section}{Sec.}{Secs.}
\Crefname{section}{Section}{Sections}
\Crefname{table}{Table}{Tables}
\crefname{table}{Tab.}{Tabs.}

\title{Towards Fine-Grained Vision-Language  Alignment for \\ Few-Shot Anomaly Detection}

\author{Yuanting Fan\textsuperscript{\rm 1,*}, Jun Liu\textsuperscript{\rm 1,*}, Xiaochen Chen\textsuperscript{\rm 1}, Bin-Bin Gao\textsuperscript{\rm 1},\\ Jian Li\textsuperscript{\rm 1}, Yong Liu\textsuperscript{\rm 1}, Jinlong Peng\textsuperscript{\rm 1}, Chengjie Wang\textsuperscript{\rm 1,\dag}\\ 
\textsuperscript{\rm 1}Tencent Youtu Lab, Shenzhen, China\\
{\tt\small \{retofan,juliusliu,husonchen,danylgao,swordli,choasliu,jeromepeng,jasoncjwang\}@tencent.com}}

\begin{document}
\maketitle
\renewcommand{\thefootnote}{\fnsymbol{footnote}} 
\footnotetext[1]{Equal contribution.}
\footnotetext[2]{Corresponding author.}





\begin{abstract}
Few-shot anomaly detection~(FSAD) methods identify anomalous regions with few known normal samples. 
Most existing methods rely on the generalization ability of pre-trained vision-language models~(VLMs) to recognize potentially anomalous regions through feature similarity between text descriptions and images. 
However, due to the lack of detailed textual descriptions, these methods can only pre-define image-level descriptions to match each visual patch token to identify potential anomalous regions, which leads to the semantic misalignment between image descriptions and patch-level visual anomalies, achieving sub-optimal localization performance. 
To address the above issues, we propose the Multi-Level Fine-Grained Semantic Caption~(MFSC) to provide multi-level and fine-grained textual descriptions for existing anomaly detection datasets with automatic construction pipeline. 
Based on the MFSC, we propose a novel framework named FineGrainedAD to improve anomaly localization performance, which consists of two components: Multi-Level Learnable Prompt~(MLLP) and Multi-Level Semantic Alignment~(MLSA). MLLP introduces fine-grained semantics into multi-level learnable prompts through automatic replacement and concatenation mechanism, while MLSA designs region aggregation strategy and multi-level alignment training to facilitate learnable prompts better align with corresponding visual regions. 
Experiments demonstrate that the proposed FineGrainedAD achieves superior overall performance in few-shot settings on MVTec-AD and VisA datasets. 
\end{abstract}

\section{Introduction}
Anomaly Detection aims to recognize and localize the anomalous regions in industrial or medical scenarios, which plays an important role in quality monitoring~\cite{bergmann2019mvtec, zou2022visa,wang2025softpatch+,li2022towards,wang2025m3dm,sun2025unseen,carrera2019online} and prognostic analysis~\cite{huang2024adapting}. Traditional methods~\cite{defard2021padim, roth2022towards,huang2022regad} optimize perceptual models based on the enormous normal images, while achieving superior performance, have limited generalization ability to unknown scenarios which lack accessible normal samples~\cite{defard2021padim, huang2022regad}. 
Due to the above issues, few-shot anomaly detection~(FSAD) methods~\cite{jeong2023winclip,li2024promptad,gu2024anomalygpt,gu2025univad, tao2025kagprompt, yao2024resad} that require only several normal images to achieve satisfactory performance have attracted increasing attention. 

\begin{figure}[t]
\centering
\includegraphics[width=1.0\columnwidth]{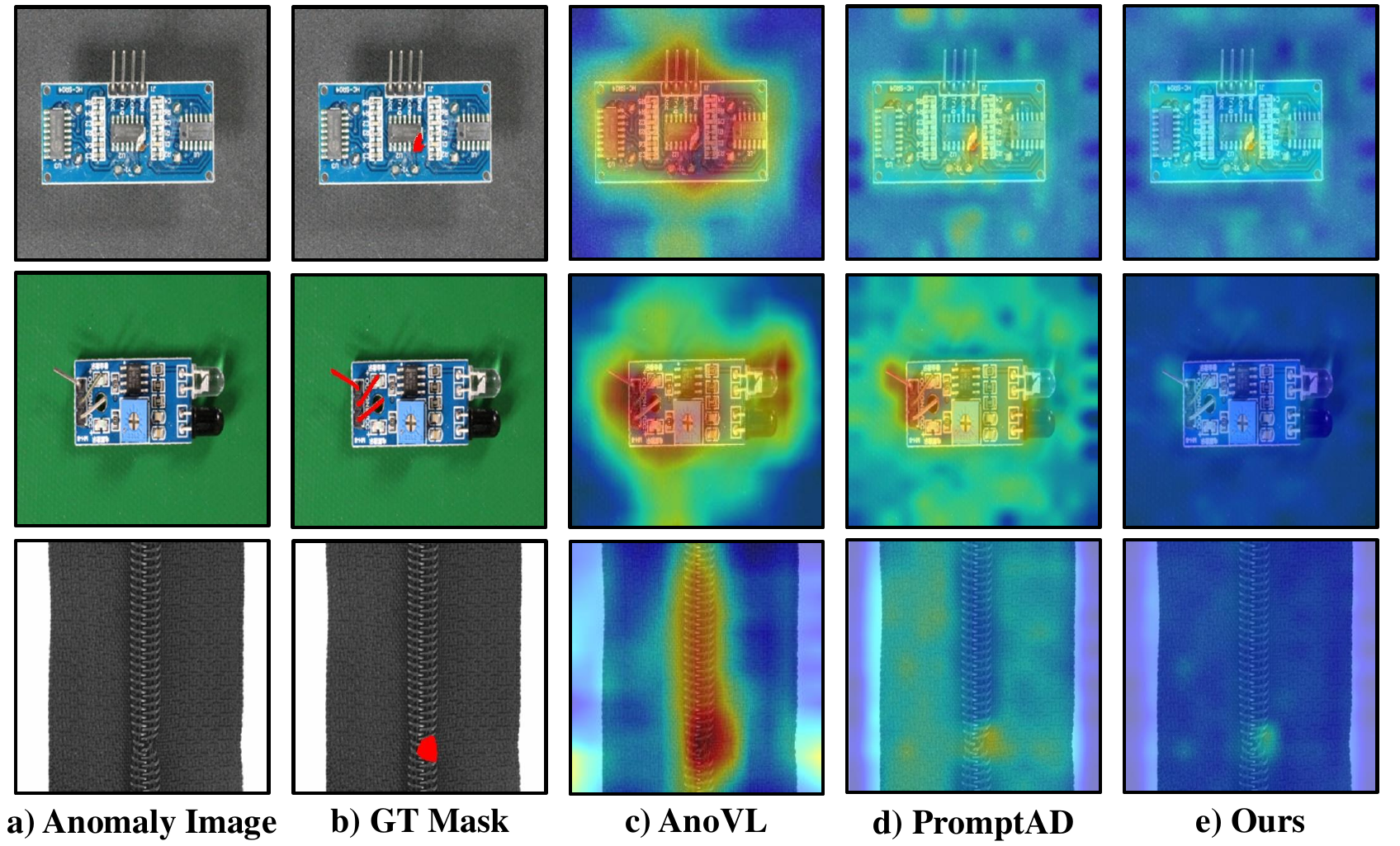}
\caption{The perceptional visualization of different methods. Previous coarse-grained handcrafted~(e.g., AnoVL) and learnable~(e.g., PromptAD) image-level prompts leads to higher activation values among normal regions, suggesting semantic misalignment between image-level prompts and patch-level visual features.}
\label{fig:teaser}
\end{figure}

Pre-trained Vision-Language models~(\eg, CLIP~\cite{radford2021clip}) have recently demonstrated strong zero/few-shot generalization ability in various downstream scenarios~\cite{jeong2023winclip, li2024promptad,jiang2024mmad,gui2024few,gao2024learning,zhang2024learning,liu2024unsupervised,gao2025adaptclip,cao2024adaclip,cai2025towards}. Based on the above foundation models, many works such as WinCLIP~\cite{jeong2023winclip}, AnomalyGPT~\cite{gu2024anomalygpt}, AnoVL~\cite{deng2023anovl}, \etc. achieve anomaly detection through calculating the cosine similarity between all image visual patches and manually crafted image-level text descriptions. 
Although these works improve the performance of FSAD, they still suffer from the problem that manually crafted image-level text descriptions~(\eg, a photo of damaged bottle) fail to accurately match the diverse anomaly visual regions across different object categories. 
Noticing this phenomenon, inspired by CoOp~\cite{zhou2022learning}, PromptAD~\cite{li2024promptad} proposes a simple and effective Semantic Concatenation~(SC) mechanism to obtain learnable prompts, and then designs explicit loss to constrain the feature distribution of normal and abnormal learnable image-level prompts, further improving the FSAD performance. 
Despite introducing learnable prompts to enhance performance, the above methods still stay at a image-level view and fail to dive into the fine-grained perception~(\eg, pins, resistors in PCB), thus limiting the overall anomaly localization capacity. 
As shown in Fig.~\ref{fig:teaser}, even PromptAD~\cite{li2024promptad} incorporates learnable prompts, the comparison between image-level prompts and patch-level visual components~(\eg, LED on PCB) results in normal regions with higher anomaly activation values.
Although UniVAD~\cite{gu2025univad} introduced an external segmentation model~(\ie, Grounded-SAM~\cite{ren2024groundedsam}) to improve the visual granularity of subsequent anomaly detection, the external models limited its further application in situations with limited computing resources due to the number of model parameters and the need for additional preprocessing. 
In this paper, our goal is to provide a universal and straight industrial anomaly detection pipeline through fine-grained alignment strategy between multi-level learnable prompts and corresponding visual components, which achieves good overall performance (\ie, localization performance and inference speed) under limited computational resources, thereby improving the effectiveness of anomaly detection in real-time applications.

Due to the utilization of only image-level prompts, existing methods suffer from the following two drawbacks: 
1) Existing AD datasets lack fine-grained textual descriptions, resulting in the construction of only image-level prompts and thus failing to fully leverage the fine-grained semantic extraction capabilities of VLMs. 
2) Using image-level prompts to calculate pixel-level anomaly scores with all visual patches fails to consider that different visual patches belong to different visual components rather than single image-level prompt, thus causing the semantic misalignment between image-level prompt features and patch-level visual token features.  

To alleviate the aforementioned issues, we first propose ulti-Level Fine-Grained Semantic Caption~(MFSC), which constructs a multi-level image description architecture~(\ie, image-level, foreground-level, and component-level) to describe the semantic information of a normal image from coarse to fine. Compared to previous simple image-level textual descriptions, it provides finer-grained textual information to better utilize pre-trained VLM's perceptional capabilities. 
Through elaborating suitable system prompts to facilitate advanced MLLMs~(\eg, GPT-4o) to generate descriptions of the visual components and corresponding attributes~(\eg, color, direction) automatically follow the above description architecture, we finally obtained MFSC suited for anomaly detection and localization tasks.
Based on the proposed MFSC, we propose a novel FSAD framework named FineGrainedAD to seamlessly solve the above two problems. FineGrainedAD consists of two indispensable components: Multi-Level Learnable Prompt~(MLLP) and Multi-Level Semantic Alignment~(MLSA). Firstly, MLLP further decomposes previous image-level prompts into multi-level prompts for fine-grained alignment prompt learning capability through automatic replacement and concatenation mechanism. 
Meanwhile, MLSA designs language-guided progressive region aggregation strategy to guide the aggregation of visual patches from the same visual component based on the MLLP, then utilizes multiple optimization objectives to explicitly align multi-level learnable prompts and corresponding visual representations during the training stage. During inference, MLSA assigns the most appropriate prompt features dynamically to each visual patch to recognize patch-level anomalies, thus achieving fine-grained anomaly localization. 
As illustrated in Fig.~\ref{fig:teaser}~c),~d)~and~e), existing image-level prompts fail to recognize the diverse types of anomalies within different visual components, while the proposed FineGrainedAD with MFSC is capable of perceiving and localizing the above anomalies accurately.  

Our contribution can be summarized as follows:
\begin{itemize}
    \item We propose the Multi-Level Fine-Grained Semantic Caption~(MFSC) and the corresponding automatic construction pipeline to promote the development of anomaly detection field from image-level to component-level perceptional granularity. 
    \item Multi-Level Learnable Prompt~(MLLP) decomposes and enriches previous image-level prompts into multi-level learnable prompts through automatic replacement and concatenation mechanism, better utilizing fine-grained semantic perception within pre-trained VLMs. 
    \item Multi-Level Semantic Alignment~(MLSA) utilize multi-level alignment training and dynamic token-wise inference mechanism between multi-level learnable prompts and corresponding visual patches to facilitate patch-level anomaly localization performance. 
    \item Extensive experiments on multiple anomaly detection datasets demonstrate that the proposed FineGrainedAD achieves state-of-the-art localization performance and inference efficiency without auxiliary training images under few-shot AD settings. 
\end{itemize}

\section{Related Work}

\noindent\textbf{Unsupervised Anomaly Detection.}
Due to the scarcity of abnormal samples, most of the existing methods rely on unsupervised learning strategy and can be categorized into following three paradigms: knowledge distillation paradigm, feature embedding paradigm, and reconstruction-based paradigm. The knowledge distillation paradigm~\cite{batzner2024efficientad, gu2023remembering, wang2021student} constrain the student model to learn only the features of the teacher model in normal samples, thus enabling anomaly detection based on the difference between student and teacher model. The feature embedding paradigm~\cite{bae2023pni, cohen2020sub, defard2021padim, liu2023simplenet} extract visual features using powerful network to perform anomaly detection. The reconstruction-based paradigm~\cite{zavrtanik2021draem, chen2024glass, zhang2023diffad} assumes that the generative model obtained by learning only normal samples can reconstruct anomaly samples into normal samples, thus accomplishing anomaly detection based on the difference between the reconstructed and original images.

\noindent\textbf{Vision-Language Models.}
In recent years, multimodal large language models (MLLMs) \cite{liu2024llava, wu2023nextgpt, bai2023qwen} have developed rapidly, especially vision-language models~(VLMs)~\cite{radford2021clip, cherti2023openclip, fan2024adadiffsr}, which have achieved significant progress in various downstream tasks~\cite{jeong2023winclip,li2024promptad,deng2023anovl,jin2025dual,gao2024metauas}. CLIP is one of the most influential VLMs, which designs two structurally similar encoders to extract features from text and images and aligns text and visual features on large-scale image-text datasets using contrastive learning optimization. With tailored prompts, CLIP achieves remarkable zero-shot image classification performance. To reduce the cost of curating appropriate prompts, inspired by prompt learning~\cite{jiang2020can} in Natural Language Processing~(NLP), recent methods like CoOp~\cite{zhou2022learning} and PromptAD~\cite{li2024promptad} using learnable rather than manually designed prompts, which only requires minimal training cost while achieving superior performance in downstream works. However, the above methods are limited to coarse-level prompts and cannot fully utilize the fine-grained perception capability within pre-trained VLMs.

\noindent\textbf{Few-shot Anomaly Detection.}
Few-shot anomaly detection~(FSAD) achieves anomaly detection and localization using only few normal images. Existing methods like SPADE \cite{cohen2020sub}, PaDiM~\cite{defard2021padim}, and DifferNet~\cite{rudolph2021same} present diverse pipeline to achieve anomaly detection, which fully utilizes latent representation of few normal images from pre-trained feature extractor. Recently, WinCLIP~\cite{jeong2023winclip} introduced CLIP and designed various handcrafted prompts to improve the performance of zero/few-shot anomaly detection. However, handcrafted prompts require tremendous cost of prompt engineering. Inspired by the learnable prompts from CoOp~\cite{zhou2022learning}, PromptAD utilizes Semantic Concatenation to construct learnable normal and abnormal prompts using fewer abnormal state words, then proposes Explicit Anomaly Margin~(EAM) to optimize the prompt feature space, thus achieving impressive performance. 
KAG-Prompt~\cite{tao2025kagprompt} utilizes multiple convolutional kernels of different sizes to extract cross-layer image visual relationships and proposes a graph learning approach to update visual prompts for anomaly detection. 
UniVAD~\cite{gu2025univad} adopts Grounded-SAM~\cite{ren2024groundedsam} and K-Means clustering algorithm to extract segment component map from few-shot normal images and test images, respectively, and then learns the relationships between the components to improve the generalization ability of anomaly detection.

\begin{figure*}[t]
\centering
\includegraphics[width=2.0\columnwidth]{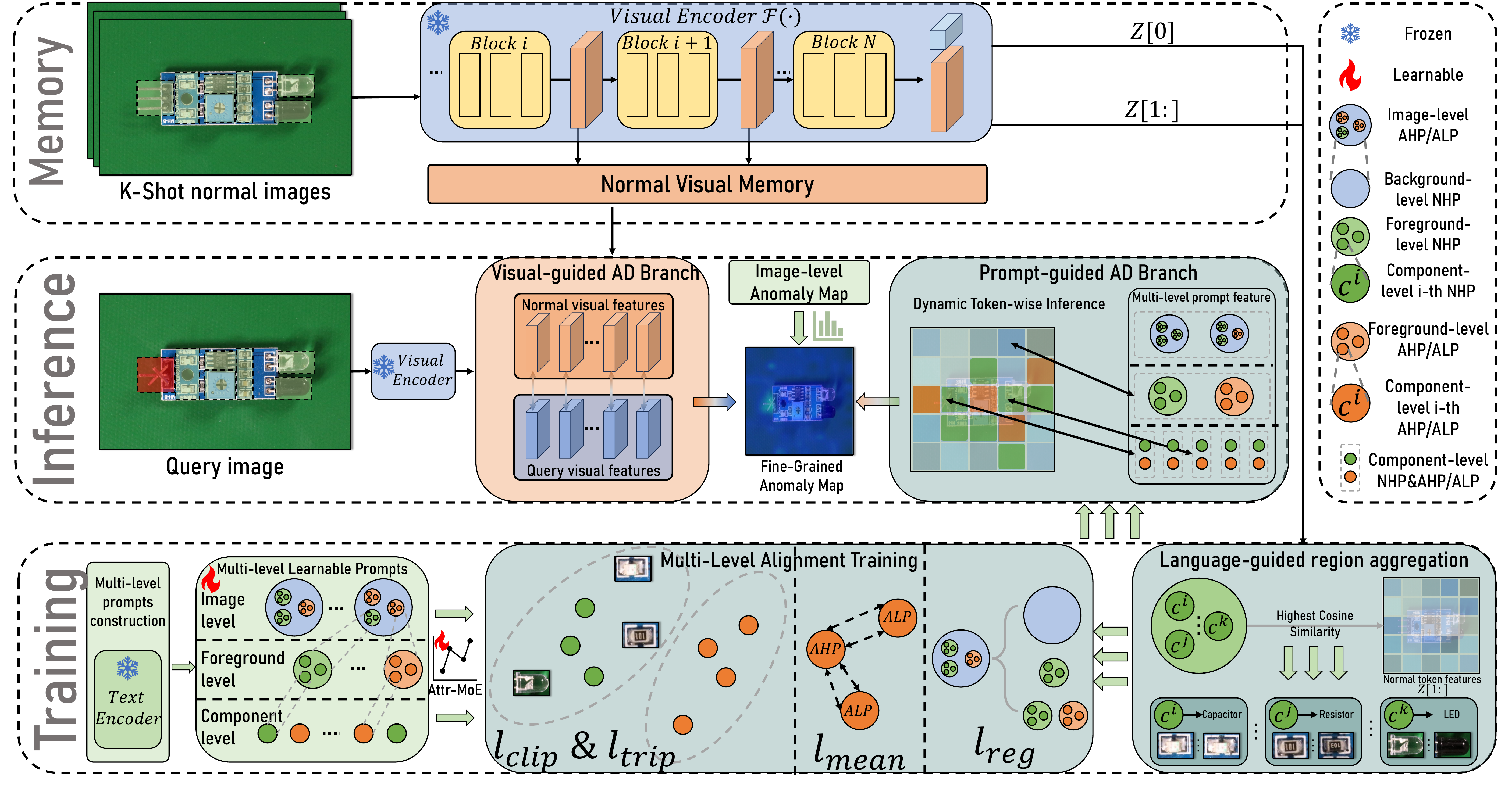}
\caption{Overview of the FineGrainedAD, which includes two branches: Vision-guided Anomaly Detection~(VAD) branch and Prompt-guided Anomaly Detection~(PAD) branch. VAD branch extracts visual features of few-shot normal images using CLIP, then compares them with the query image features to obtain the anomalous regions. 
PAD branch first utilize language-guided region aggregation to obtain matching relationship between multi-level prompts and corresponding visual regions, then optimizes the feature space of multi-level learnable prompts through multi-level alignment training, further adopts dynamic token-wise inference mechanism to assign appropriate prompt to each visual patch during inference, achieving accurate perception within each visual component.}
\label{fig:overview}
\end{figure*}

\section{Methodology}

\subsection{Motivation and Overview}

Existing prompt learning-based FSAD methods suffer from the following drawbacks: 
the designed prompts~(handcrafted or learnable) are only limited to the image-level scope~(e.g., the photo of [normal/abnormal state words] object), although some methods~\cite{gu2024filo, deng2023anovl} introduce position or domain information to improve the prompt granularity, they still fail to penetrate to the component-level perceptual granularity and only achieve sub-optimal results. Meanwhile, utilizing the same image-level prompt to compute anomaly scores with diverse visual components also leads to serious semantic misalignment. 
Therefore, we first constructed the image description architecture of Multi-Level Fine-Grained Semantic Caption~(MFSC), then automatically generated MFSC using the fine-grained perception capabilities of existing MLLMs, as shown in Sec.~\ref{subsec:mfsc}. 
Furthermore, we propose FineGrainedAD, as shown in Fig.~\ref{fig:overview}, which can be divided into two branches: Vision-guided Anomaly Detection~(VAD) and Prompt-guided Anomaly Detection~(PAD). 
Firstly, VAD stores the patch token features from particular layers of the CLIP visual encoder in normal visual memory as ${R}$ during training. During inference, we adopt same visual encoder to extract the same layer features from Query image and compare the query features ${F} \in \mathbb{R}^{h\times w\times d}$ with memory features ${R}$ to obtain the vision-guided anomaly score map ${M}_{v} \in [1, 0]^{h\times w}$: 
\begin{align}
{M}_{v, ij} &= \min_{\vec{r} \in {R}} \frac{1}{2} (1 - <\vec{f}_{ij}, \vec{r}>),
\label{eq:vad}
\end{align}
Besides, the PAD branch consists of the following two parts: Multi-Level Learnable Prompt~(MLLP) in Sec.~\ref{subsec:mllp} and Multi-Level Semantic Alignment~(MLSA) in Sec.~\ref{subsec:mlsa}.

\subsection{Multi-Level Fine-Grained Semantic Caption}
\label{subsec:mfsc}
Existing prompt learning-based FSAD methods can only utilize the foreground object name and pre-defined anomaly state words for image-level prompt construction, leading to the above mentioned semantic confusion. 
Therefore, it is critical to introduce fine-grained semantic representation to build prompts for more accurate anomaly perception and localization capability. 

\begin{figure}[!t]
\centering
\includegraphics[width=1.0\columnwidth]{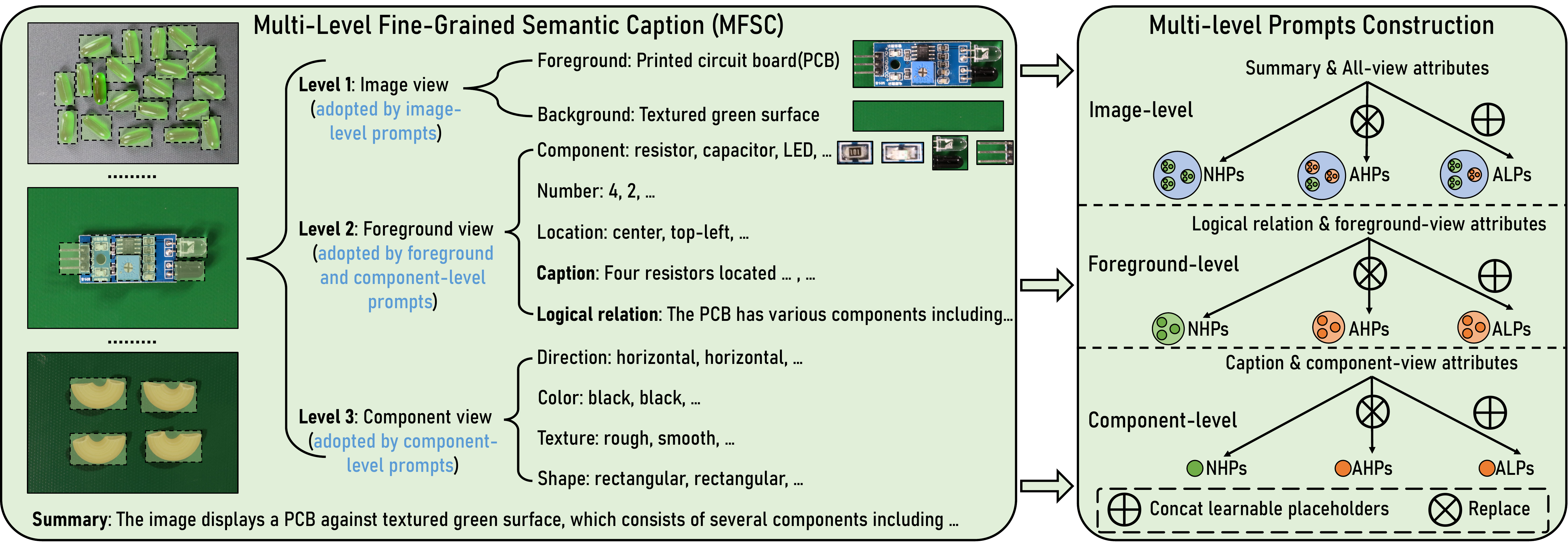}
\caption{The visualization of Multi-Level Fine-Grained Semantic Caption~(MFSC) and Multi-level Prompts Construction through replacement and concatenation mechanism.}
\label{fig:mfsc}
\end{figure}

Inspired by human visual perception, we have crafted image description architectures with three semantic granularities from coarse to fine, \ie, image view, foreground view, and component view. The above three levels of semantic granularity effectively describe an entire image from global to local details. 
As shown in Fig.~\ref{fig:mfsc}, the image view divides the image into foreground and background regions, while the foreground view focuses on the relations between different visual components in the foreground regions, such as number, position, etc. 
By combining the above attributes information, we can obtain the detailed visual component descriptions and logical relationship between different components~(\ie, \texttt{Caption} and \texttt{Logical relation} in Fig.~\ref{fig:mfsc}). 
Besides, to refine the perceptual granularity to the component-level, we add identifiable component features to the description architectures, such as color, texture, and other attributes to enrich the component descriptions. Then, we obtain the comprehensive image-level summary~(i.e., \texttt{Summary} in Fig.~\ref{fig:mfsc}) according to the above automatically generated multi-level descriptions. 
To deal with the diversity of the component attribute distribution, such as mixed colors, special shapes, \etc, we set the base and extension value for each attribute. Taking \texttt{Color} as an example, the base value is defined as the common colors such as black, white, \etc. For the mixed colors or multiple potential colors, we adopt \textit{and}, \textit{or}, \textit{with} to connect the base color values, thus describing the visual component more accurately and flexibly.

\begin{figure}[t]
\centering
\includegraphics[width=1.0\columnwidth]{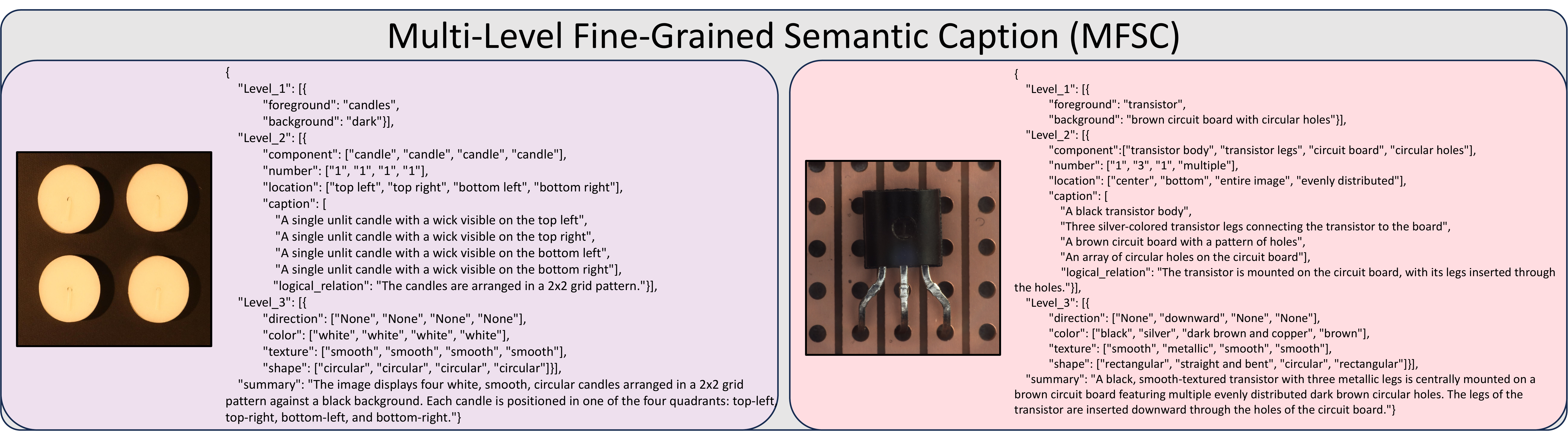}
\caption{The visual example of automatically constructed Multi-Level Fine-Grained Semantic Caption~(MFSC) on MVTec-AD and VisA datasets.}
\label{fig:MFSC_sample}
\end{figure}

Furthermore, manual annotation introduce significant workload and high degree of uncertainty, to reduce the cost of manually labeling, we curate system prompt to facilitate advanced MLLMs~(e.g., GPT-4o, Gemini-2.5) to generate Multi-Level Fine-Grained Semantic Caption~(MFSC) according to the single normal image. 
Although MLLMs are less effective as an anomaly detector, when used as a semantic attribute extractor, they can accurately recognize the visual components and corresponding attributes information. 
The above description architectures can be applied to mainstream anomaly detection datasets, such as MVTec-AD~\cite{bergmann2019mvtec},VisA~\cite{zou2022visa}, Real-IAD~\cite{wang2024realiad}, \etc. 
The proposed MFSC also significantly enhances the performance of existing VLM-based AD methods through In-Context Learning~(ICL) strategy. 
We visualize the example of MFSC automatically curated for the MVTec-AD~\cite{bergmann2019mvtec} and VisA~\cite{zou2022visa} datasets, as shown in Fig.~\ref{fig:MFSC_sample}. 
More details of MFSC are provided in the supplementary materials.

\subsection{Multi-level Learnable Prompts}
\label{subsec:mllp}
Based on the MFSC, we can seamlessly migrate the semantic perception capability from the image-level~(existing methods) to patch-level, thus better utilizing the prior within VLMs to accomplish the fine-grained anomaly detection. 
Specifically, we construct prompts at the same three levels of granularity: image-level, foreground-level, and component-level. 
Each level of prompts can be categorized into three types: normal handcrafted prompts~(NHPs), abnormal handcrafted prompts~(AHPs), and abnormal learnable prompts~(ALPs). 
The reason for this design~(\ie, only introduce extra learnable prompts for abnormal representation) is that while normal samples can be described by relatively deterministic text descriptions, anomaly samples are more diverse and unknown, thus we only design learnable prompts for anomaly samples to accommodate diverse distribution. 
Meanwhile, to improve the robustness and flexibility of NHPs, we insert learnable placeholders to align them with few-shot normal samples. 

Firstly, for the image-level prompts, we use the \texttt{Summary} from MFSC as the NHPs. Since \texttt{Summary} contains multiple attribute descriptions, we then utilize replacement mechanism to construct the corresponding AHPs. Specifically, we replace the attribute values by adding negative descriptions or randomly selecting other values from the corresponding attribute list. 
For the image-level ALPs, we replace the multiple attribute values with learnable placeholders, thus realizing the ability to perceive and accommodate diverse anomalous distribution. 
Secondly, we use the \texttt{Logical relation} to build the foreground-level NHPs, which mainly describe the logical relationships within foreground regions such as the number of components and their relative spatial location. 
Similar to image-level prompts, we construct foreground-level AHPs and ALPs through concatenate or replace mechanisms. 
Note that we also design background-level NHPs using \texttt{Background} for subsequent anomaly localization. 
Finally, for the component-level prompts, we utilize the \texttt{Caption} as NHPs. Then we adopt similar replacement and concatenation mechanisms to construct AHPs and ALPs. 
It is worth noting that since different attributes have different importance levels during the perception process, to dynamically adapt to diverse anomalous distribution, we set learnable parameters between zero and one for each learnable placeholder of component-level prompts~(i.e., Attr-MoE in Fig.~\ref{fig:overview}). 
More details of MLLP are provided in the supplementary materials.

\subsection{Multi-Level Semantic Alignment}
\label{subsec:mlsa}
After the automatic construction of multi-level learnable prompts, we are able to use them in alignment training with few-shot normal images. Existing methods typically use learnable image-level prompts and all visual patch tokens for alignment training, then compute anomaly score between image-level prompts and each visual token during inference, which obviously leads to serious semantic misalignment and sub-optimal learning ability. 
To address the above issues, we first introduce a language-guided progressive region aggregation strategy, which divides images into multiple regions based on the semantic differences, and then allows learnable prompts to pay more attention to their corresponding visual regions during alignment training. 
Meanwhile, we also adopt dynamic token-wise inference strategy to compute anomaly scores for learnable prompts and their corresponding visual regions, achieving fine-grained anomaly localization. 

\noindent\textbf{Language-guided progressive region aggregation.} Specifically, for an input normal image, we adopt the CLIP visual encoder to provide the original visual tokens $\mathbf{Z} = {\left \{ \vec{z_{i}} \right \}}^{T}_{i=1}$, where $T$ is the number of visual patch tokens. To split the image into different regions, we modify DPC-KNN~\cite{du2016dpcknn}, a k-nearest-neighbor-based density peaks clustering algorithm, to aggregate the visual tokens with same semantic categories. 
As shown in Fig.~\ref{fig:match_map}, adopting DPC-KNN directly for clustering is less effective, so we have made four modifications:
\begin{itemize}[leftmargin=1em]
    \item \textbf{Initial center selection}: Unlike DPC-KNN uses local density and relative distance, we introduce multi-level semantics for initial center selection. By calculating the visual token with the highest cosine similarity to the multi-level prompt features as the component region cluster center, we avoid performance degradation due to the initial center selection. 
    \item \textbf{Reference dimension}: Directly using visual token features for clustering, the obtained regions may not match the corresponding semantic prompts~(\ie, the token features reflect low-level texture and are less relevant to semantics). To mitigate this problem, we use the multi-dimensional cosine similarity between each visual token and multi-level prompt features as the reference dimension, thus better utilizing semantics to guide the process of region aggregation. 
    \item \textbf{Two-stage clustering}: First, we perform two classes clustering, using foreground and background-level prompts as semantic information to guide the clustering process. Then, for the obtained foreground visual tokens, we adopt component-level prompts and finally obtain the mapping relationship between visual regions and multiple component prompts. Compared to one stage, the two-stage clustering enables more accurate recognition of component visual regions. 
    \item \textbf{Higher resolution}: We found that utilizing the native resolution of the pre-trained CLIP for clustering is less effective, which is due to the information degradation after resizing original image to a smaller resolution. Therefore, we use larger resolution and feed image into CLIP after slice, then perform clustering on the higher resolution features, thus achieving better performance. Note that we only adopt higher resolution patch features during clustering process. 
\end{itemize}

\noindent\textbf{Multi-Level Alignment training.}  
After obtaining the multi-level learnable prompts constructed by MLLP and their corresponding visual regions, we can establish explicit alignment relationships to adapt each level of learnable prompts to more specific visual representations. 
Specifically, we adopt the four optimization objectives in multi-level learnable prompts for alignment training with few-shot normal images. 
Firstly, to make the normal images and visual components closer to corresponding NHPs and away from AHPs/ALPs in the feature space, follow~\cite{li2024promptad}, we utilize the CLIP Loss $\ell_{clip}$ and Triplet loss $\ell_{trip}$ in three levels of leranable prompts. Meanwhile, we utilize Mean Loss $\ell_{mean}$ between multi-level AHPs and ALPs to stabilize the learning process. As shown in the following equation:
\begin{equation}
\begin{aligned}
\ell _{clip} = -\log & \frac{\exp(<\vec{z}, \vec{p}_{n}>)}{\exp(<\vec{z}, \vec{p}_{n}>) +\sum_{\vec{p} \in \Lambda }  \exp(<\vec{z}, \vec{p}>)}, \\
\ell _{trip} &= \max\left \{ d(\vec{z}, \vec{\bar{p}}_{n} )  - d(\vec{z}, \vec{\bar{p}_{a}}) + \varepsilon ,0  \right \}, \\
&\ell _{mean} = {{\left \|   \frac{\vec{p}_{a}^{h}}{ || \vec{p}_{a}^{h}  || }_{2} - \frac{\vec{p}_{a}^{l}}{ || \vec{p}_{a}^{l}  || }_{2}  \right \|}^{2} }.
\end{aligned}
\end{equation}

For $\ell _{clip}$, $\vec{z}$ represents normal visual token features, $p_{n}$ denotes NHPs features, $<\cdot>$ is the cosine similarity, and $\Lambda$ is the union of AHPs and ALPs features. Then, for the $\ell _{trip}$, $\vec{\bar{p}}_{a}$ and $\vec{\bar{p}}_{n}$ denote the average AHPs/ALPs and NHPs features, respectively. The $\varepsilon$ means the margin between $\vec{\bar{p}}_{n}$ and $\vec{\bar{p}}_{a}$, which is fixed to $1$, and the $d(\cdot)$ denotes the euclidean distance. And for the $\ell _{mean}$, $\vec{p}_{a}^{h}$ and $\vec{p}_{a}^{l}$ denotes AHP and ALP features, respectively. 
Similar to the MLLP, the optimization granularity for $\ell_{clip}$, $\ell_{trip}$, and $\ell_{mean}$ can be divided into three levels, image-level, foreground-level, and component-level. 
Benefiting from visual regions obtained by the above region aggregation strategy, we can assign larger learning weights $\gamma$ to the corresponding visual tokens for three-level learnable prompts to better perceive the anomalous distribution.

Besides, inspired by InstantStyle~\cite{wang2024instantstyle}, different image components can be decoupled by subtracting the corresponding CLIP features. We design a novel regularization loss $\ell_{reg}$ to constrain the learning process of image- and component-level abnormal learnable prompts as:
\begin{equation}
\begin{split}
\ell_{reg} = \lambda_{reg} \left \|(\vec{p}_{a, img} - \sum_{i = 1}^{N_{c}}\vec{p}_{a, c_{i}} ) - \vec{p}_{b} \right \|,
\end{split}
\end{equation}
where $\vec{p}_{b}$ and $\vec{p}_{a, img}$ denote the background-level prompt features and image-level ALP features, respectively. The $\vec{p}_{a, c_{i}}$ means the component-level ALP features, $N_c$ is the number of components, and $\lambda_{reg}$ represents the adjustable coefficient. More details are provided in the ablation studies in the main paper.

\noindent\textbf{The PAD branch of FineGrainedAD.} 
After alignment training with few-shot normal images, the multi-level learnable prompts can support PAD branch for fine-grained anomaly detection. 
Specifically, PAD calculates token-wise anomaly score by retrieving the most appropriate learnable prompts for each visual patch token, thus achieving fine-grained anomaly detection. 
Since only normal images are accessible during training, there are significant domain gaps in directly using abnormal learnable prompt features to retrieve visual tokens of query image during inference, which leads to inaccurate prompt-visual regions matching relationship. Therefore, we utilize a Query Former to extract intrinsic representation from normal and abnormal multi-level prompt features to enhance the AHPs features for more precise retrieval.
First, we randomly initialize multi-level learnable intrinsic representation, then feed it as Query and normal/abnormal multi-level prompt features as Key, Value into the Query Former. The Query Former consists of two parallel Cross Attention Layers, one for extracting normal representation and the other for abnormal representation. Through single projection layer, we fuse the two features into multi-level intrinsic representation. During training, we maximize cosine similarity between intrinsic features and corresponding normal/abnormal prompt features to optimize Query Former.

Then, we calculate the many-to-many cosine similarity between multi-level intrinsic representations and visual token features of query image, \ie, we assign the multi-level prompt with the largest cosine similarity to each visual token and compute the token-wise anomaly score based on the corresponding normal and abnormal prompt features. As shown in the following equation: 
\begin{align}
{s(\vec{z}, \vec{p})} &= \frac{\exp(<\vec{z}, \vec{{p}}_{n}>)}{\exp(<\vec{z}, \vec{{p}}_{n}>) + \exp(<\vec{z}, \vec{\bar{p}}_{a}>)},\\
{\hat{M}}_{p,ij}   &=  s(\vec{z}_{ij} ,{\rm{argmax}}_{\vec{{p}}_{k} \in [0, N_c]}(<\vec{z}_{ij}, \vec{p}_{k}>)),
\end{align}
where $s(\vec{z}, \vec{p})$ denotes anomaly score between visual token feature $\vec{z} \in \mathbb{R}^{h\times w\times d}$ and corresponding prompt feature $\vec{p}$~(including normal prompt features $\vec{{p}}_{n}$ and average abnormal prompt features $\vec{\bar{p}}_{a}$). $N_c$ represents the number of total multi-level prompt features. 
$N_{ab}$ denotes the number of ALPs for each prompt features, and $\vec{\bar{p}}_{a} = \frac{\sum_{i=1}^{N} \vec{p}_{a, i} }{N_{ab}}$, $\vec{p}_{a, i}$ means the $i$-th abnormal prompt feature within single intrinsic representation. 
${\hat{M}}_{p} \in [1, 0]^{h\times w}$ represents prompt-guided pixel anomaly score. 
Finally, to further refine the anomaly scores, we utilize the image-level learnable prompt features to calculate the image-level anomaly distribution map, then re-weight the ${\hat{M}}_{p}$ and obtain the final pixel anomaly score map ${{M}}_{pix}$:
\begin{align}
{{M}}_{p,ij} = & \frac{\textrm{softmax} (s(\vec{{z}}_{ij}, \vec{{p}}_{img}))}{\sum_{k = 1}^{T}  \textrm{softmax} (s(\vec{z}_{k}, \vec{p}_{img} ))}  \odot  {\hat{M}}_{p,ij},\\
{{M}}_{pix} & = 1.0 /  (1.0 /{M}_{v} + 1.0 / {M}_{p}),
\end{align}
where $\vec{z} \in \mathbb{R}^{h\times w\times d}$ and $\vec{p}_{img}$ denote visual token features and image-level abnormal prompt features, respectively. 
${M}_{p} \in [1, 0]^{h\times w}$ represents the modified prompt-guided pixel anomaly score.

\begin{table*}[!t]
\small
\centering
\scalebox{0.8}{
\begin{tabular}{lccccccccc}
\hline
\multirow{2}{*}{Method} & \multirow{2}{*}{Backbone(\#Params)} & \multirow{2}{*}{Auxiliary} & \multicolumn{3}{c}{MVTec-AD~\cite{bergmann2019mvtec}} & \multicolumn{3}{c}{VisA~\cite{zou2022visa}} & \multirow{2}{*}{time(ms)} \\ \cline{4-6} \cline{7-9} 
 & &  & 1-shot  & 2-shot & 4-shot & 1-shot & 2-shot & 4-shot & \\ 
 \hline
 RegAD~\cite{huang2022regad}& ResNet-18(12M) & \ding{51} & -/- & 85.50/94.60 & 89.23/95.84 & -/- & -/- & -/- & 51.3\\
 AnomalyGPT~\cite{gu2024anomalygpt}& ImageBind-H+Vicuna(8B) & \ding{51} & 94.22/95.45 & 94.38/95.82 & 96.62/96.11 & 89.96/\textcolor{blue}{{96.39}} & 90.34/96.78 & 90.15/97.11 & 3595.4\\
 InCTRL~\cite{zhu2024inctrl}& CLIP-B(208M) & \ding{51} & -/- & 94.15/- & 94.68/- & -/- & 86.02/- & 87.89/- & 21.2\\ 
 KAG-Prompt~\cite{tao2025kagprompt}& ImageBind-H+Vicuna(8B) & \ding{51} & 92.33/\textcolor{red}{95.83} & 93.47/\textcolor{red}{96.33} & 94.03/\textcolor{blue}{{96.69}} & 91.37/96.31 & 92.11/\textcolor{red}{97.01} & 92.41/\textcolor{blue}{{97.43}} & 3595.4\\ 
 ResAD~\cite{yao2024resad}& ImageBind-H(632M) & \ding{51} & -/- & 90.37/95.22 & 91.00/96.00 & -/- & 88.73/96.12 & 89.40/96.80 & 76.3\\ 
 \hline
 PaDiM~\cite{defard2021padim}& ResNet-18(12M) & \ding{55} & 76.60/89.30 & 78.91/91.32 & 80.44/92.61 & 62.82/89.94 & 67.41/92.01 & 72.81/93.23 & 68.5\\
 SPADE~\cite{cohen2020sub}& WideResNet-50(69M) & \ding{55} & 81.01/91.20 & 82.92/92.01 & 84.84/92.71 & 79.52/95.63 & 80.72/96.21 & 81.78/96.61  & 63.1\\
 PatchCore~\cite{roth2022towards}& WideResNet-50(69M) & \ding{55} & 83.41/92.00 & 86.32/93.34 & 88.88/94.31 & 79.95/95.40 & 81.64/96.12 & 85.37/96.81 & 73.2\\
 WinCLIP+~\cite{jeong2023winclip}& CLIP-B(208M) & \ding{55} & 93.70/93.60 & 93.70/93.80 & 95.30/94.20 & 83.82/94.73 & 83.41/95.15 & 84.11/95.25 & 23.3\\
 PromptAD~\cite{li2024promptad}& CLIP-B(208M) & \ding{55} & 89.87/95.11 & 93.14/95.46 & 94.17/95.73 & 83.62/96.25 & 85.70/96.80 & 86.02/96.96 & 23.3\\ 
 \rowcolor{cyan!10}\textbf{FineGrainedAD}& CLIP-B(208M) & \ding{55} & 91.04/\textcolor{blue}{{95.49}} & 93.66/\textcolor{blue}{{96.08}} & 94.51/\textcolor{red}{96.71} & 84.16/\textcolor{red}{96.42} & 86.16/\textcolor{blue}{{96.97}} & 86.95/\textcolor{red}{97.54} & 23.3\\ 
 \hline
\end{tabular}
} 
\caption{Quantitative comparisons of image/pixel anomaly detection in AUROC on MVTec-AD and VisA datasets. The best and second best pixel performances are in \textcolor{red}{red} and \textcolor{blue}{{blue}} colors, respectively. Auxiliary denotes whether additional training data are used.}
\label{tab:fs_metrics}
\end{table*}

\begin{table*}[!t]
\begin{center}
\begin{minipage}{0.48\textwidth}
\centering
\tiny
\resizebox{\linewidth}{!}{
\begin{tabular}{lcccc}
\hline
Method & Public & Setting & image & pixel \\
\hline
 \rowcolor{cyan!10}\textbf{Ours} & - & 1-shot & 84.2 & 96.4\\
 \rowcolor{cyan!10}\textbf{Ours} & - & 4-shot & 87.0 & 97.5 \\ \hline
 WinCLIP+~\cite{jeong2023winclip} & \textit{CVPR 2023} & 8-shot & 86.0 & \textbf{95.3} \\
 PaDiM~\cite{defard2021padim} & \textit{CVPR 2022} & 16-shot & 78.6 & \textbf{95.8} \\ \hline
 DSR~\cite{zavrtanik2022dsr} & \textit{ECCV 2022} & full-shot & 88.0 & \textbf{84.3} \\
 D3AD~\cite{tebbe2024d3ad} & \textit{CVPRW 2024} & full-shot & 96.0 & \underline{97.9} \\  
 GLASS~\cite{chen2024glass} &  \textit{ECCV 2024} & full-shot & 98.8 & \underline{98.8} \\ \hline
\end{tabular}
}
\caption{Comparisons with many-shot methods in AUROC on VisA. The performance below and above our 1-shot methods are in \textbf{bold} and \underline{underlined}, respectively.}
\label{tab:ms_metrics}
\end{minipage}
\hfill
\begin{minipage}{0.46\textwidth}
\centering
\tiny
\resizebox{\linewidth}{!}{
\begin{tabular}{cccccc}
\hline
\multicolumn{4}{c}{Components}  & \multicolumn{2}{c}{VisA} \\
 \cline{1-4} \cline{5-6}
 $\ell_{clip}$ & $\ell_{trip}$ & $\ell_{mean}$ & $\ell_{reg}$        & image       & pixel      \\ \hline
 \ding{55} & \ding{55} & \ding{55} & \ding{55} & 82.88 & 95.15 \\
 \ding{51} & \ding{55} & \ding{55} & \ding{55} & 84.25 & 96.62 \\
 \ding{51} & \ding{51} & \ding{55} & \ding{55} & 85.93 & 96.91 \\
 \ding{51} & \ding{51} & \ding{51} & \ding{55} & 86.41 & 97.26\\
 \rowcolor{cyan!10}\ding{51} & \ding{51} & \ding{51} & \ding{51} & 86.95 & 97.54 \\    
\hline
\end{tabular}
}
\caption{Image/ pixel-level AUROC of different settings on optimization objectives. The first row denotes only using handcrafted prompts in PAD and entire VAD branch.}
\label{tab:loss_abl}
\end{minipage}
\end{center}
\end{table*}

For image-level anomaly classification, we utilize the maximum value of ${M}_{pix}$ and image-level score ${S}_{i}$ to obtain the final image anomaly score ${{M}}_{img}$:
\begin{equation}
\begin{aligned}
{S}_{i} &= s(\vec{z}_{cls}, \vec{p}_{img}),\\
{M}_{img} = 1.0 / & (1.0 / \max {M}_{pix}  + 1.0 / {S}_{i}),
\end{aligned}  
\end{equation}
where $\vec{z}_{cls}$ denotes the class token features of the query image. Following previous works~\cite{li2024promptad, jeong2023winclip}, we use harmonic mean to fuse ${M}_{pix}$ and ${S}_{i}$ to obtain final image-level anomaly score, which is more sensitive to smaller values.

\section{Experiments}

\subsection{Experimental Setup}
\noindent\textbf{Datasets.}
We conduct our experiments on MVTec-AD~\cite{bergmann2019mvtec} and VisA~\cite{zou2022visa} datasets. Both benchmarks have multiple anomalous types within different objects. MVTec consists of 15 categories with $700^{2}-900^{2}$ pixels, VisA contains 12 categories with roughly $1.5\mathrm{K}\times 1\mathrm{K}$ resolution. The training set contains only normal images, while the test set contains normal and abnormal images with pixel-level annotations.

\noindent\textbf{Comparative Methods and Evaluation Metrics.}
We compare the proposed FineGrainedAD with existing state-of-the-art FSAD methods under 1, 2, and 4-shot settings, which include both image- and pixel-level metrics. Following \cite{jeong2023winclip, li2024promptad}, for three conventional full-shot AD methods, including SPADE~\cite{cohen2020sub}, PaDiM~\cite{defard2021padim}, and PatchCore~\cite{roth2022towards}, we adapt their distance-based scoring mechanism to few-shot settings for fair comparisons. 
We also conduct experiments on many-shot and full-shot methods to validate the advantages of component-level anomaly detection. 
Additionally, to demonstrate the effectiveness of the proposed method on the RealIAD~\cite{wang2024realiad} dataset, we compared its performance against our baseline model PromptAD~\cite{li2024promptad}.

\noindent\textbf{Implementation details.}
For CLIP-based methods, including WinCLIP, PromptAD, InCTRL, and our FineGrainedAD, we adopt the same CLIP implementation, OpenCLIP~\cite{cherti2023openclip}, and its pre-trained backbone ViT-B/16+~\cite{dosovitskiy2020vit} on LAION-400M dataset. 
The resolution used in region aggregation is $4\times$ than CLIP input resolution($240\times240$). 
We use SGD as the optimizer and initial learning rate is set to 2e-3. 
During training, we only update the parameters of multi-level learnable prompts, Query Former, and Attr-MoE. 
Note that for the comparison experiments of different methods on the RealIAD~\cite{wang2024realiad} dataset, we only conducted testing exclusively from the C1 view to validate generalization capabilities and reduce computational cost. 
More training details of FineGrainedAD are provided in the supplementary materials.

\begin{figure*}[t]
\centering
\includegraphics[width=2.0\columnwidth]{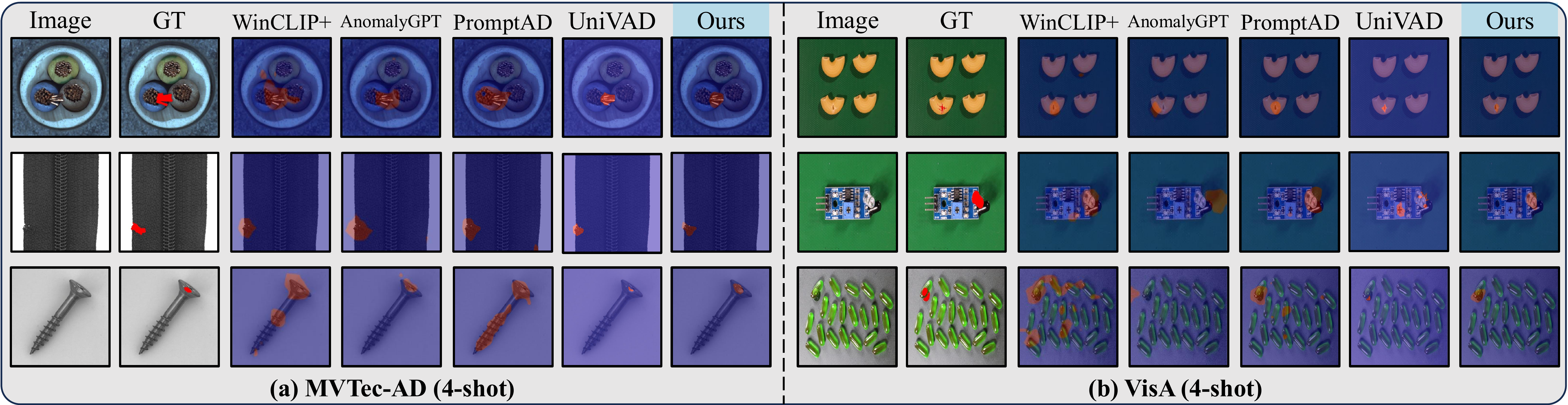}
\caption{The qualitative comparisons of 4-shot anomaly localization methods on MVTec-AD and VisA.}
\label{fig:bigpic}
\end{figure*}

\begin{table*}[t]
\begin{center}
\begin{minipage}{0.48\textwidth}
\centering
\resizebox{\linewidth}{!}{
\tiny
\begin{tabular}{cccccccc}
\hline
\multicolumn{4}{c}{Components} & \multicolumn{2}{c}{MVTec-AD} & \multicolumn{2}{c}{VisA} \\ \hline
BL   & MP     & AT   & TI    &  image       & pixel  & image       & pixel    \\ \hline
 \ding{51} & \ding{55} & \ding{55} & \ding{55}  & 94.17 & 95.73 & 86.02 & 96.96 \\
  \ding{51} & \ding{51} & \ding{55} & \ding{55} & 94.33 & 96.05 & 86.51 & 97.15 \\ 
 \ding{51} & \ding{51} & \ding{51} & \ding{55} & 94.48 & 96.38 & 86.84 & 97.43 \\
 \ding{51} & \ding{51} & \ding{55} & \ding{51} & 94.37 & 96.24 & 86.67 & 97.28 \\
 \rowcolor{cyan!10}\ding{51} & \ding{51} & \ding{51} & \ding{51} & 94.51 & 96.71 & 86.95  & 97.54 \\ \hline
\end{tabular}
}
\caption{Image- and pixel-level AUROC of different components. BL, MP, AT, and TI denote the baseline, multi-level learnable prompts, multi-level alignment training, and dynamic token-wise inference mechanism, respectively.}
\label{tab:comp_abl}
\end{minipage}
\hfill
\begin{minipage}{0.46\textwidth}
\centering
\tiny
\resizebox{\linewidth}{!}{
\begin{tabular}{lccccc}
\hline
\multicolumn{4}{c}{Backbone} & \multicolumn{2}{c}{VisA} \\
\hline
Name   & Param   & Res & Token  & image       & pixel      \\
\hline
 ViT-B/32 & 86M & 224 &  49   & 84.94 & 96.11 \\
 ViT-B/16 & 86M & 224 &  196   & 85.34 &  96.91 \\
 ViT-B/16+ & 208M & 224 &  196  & 86.83  & 97.33 \\
 \rowcolor{cyan!10}ViT-B/16+ & 208M & 240 &  225   & 86.95 & 97.54 \\ 
 ViT-L/14 & 307M & 224 &  256   & 87.11 & 97.61 \\ 
 ViT-L/14 & 307M & 336 & 576 &  87.54 & 97.88 \\ 
 ViT-H/14 & 632M & 224 & 256 & 88.12 &  97.62 \\ 
\hline
\end{tabular}
}
\caption{Image/pixel-level AUROC of different backbones. Param, Res, and Token is the parameters, resolution, and token numbers, respectively.}
\label{tab:backbone_abl}
\end{minipage}
\end{center}
\end{table*}

\subsection{Comparison Results}
\noindent\textbf{Comparisons with few-shot methods.}
We first conduct the comparisons with state-of-the-art few-shot anomaly detection methods on MVTec-AD and VisA datasets. As shown in Tab.~\ref{tab:fs_metrics}, the proposed FineGrainedAD outperforms existing methods on pixel-level AUROC in three few-shot settings without the burden of auxiliary training data~(e.g., AnomalyGPT), multiple inference~(e.g., RegAD), and extra networks with enormous parameters~(e.g., AnomalyGPT). It's worth noting that AnomalyGPT utilizes MLLM to boost the image-level anomaly classification performance. 
Specifically, compared with state-of-the-art CLIP-based methods PromptAD, FineGrainedAD achieves $0.4\%/1.0\%$ and $1.1\%/0.6\%$ improvements under the 4-shot settings of MVTec-AD and VisA, respectively. 
As shown in Fig.~\ref{fig:bigpic}, the proposed FineGrainedAD optimizes the perception granularity of FSAD methods, thus achieving finer-grained anomaly localization. 
Meanwhile, we also compare the inference times of the different methods under 4-shot setting, as shown in Tab.~\ref{tab:fs_metrics}, which shows that due to the lightweight and efficient nature of the adopted strategy, FineGrainedAD achieves a trade-off between anomaly localization performance and inference efficiency compared to other methods that require external training data and powerful perceptual models.
As shown in Tab.~\ref{tab:fs_metrics_realiad}, on datasets like RealIAD that feature more diverse distributions and higher authenticity, FineGrainedAD achieves superior prompt learning capability and anomaly localization performance compared to WinCLIP and PromptAD, further validating the proposed method's generalization capability and effectiveness of MFSC.

\noindent\textbf{Comparisons with many-shot methods.}
We also conduct comparisons with many/full-shot methods on VisA dataset. As shown in Tab.~\ref{tab:ms_metrics}, FineGrainedAD achieves better image- and pixel-level results than other many-shot AD methods, which demonstrates the robustness of FineGrainedAD under few-shot settings. Besides, FineGrainedAD is also superior to the early full-shot AD methods, further minimizing the gap between few-shot and full-shot AD methods.

\subsection{Ablation Study and Discussions}
We verify the impact of different modules under 4-shot settings on MVTec-AD and VisA datasets, which mainly include optimization objectives and other network components. Meanwhile, we also analyze the effect of alignment training and pre-trained backbones with different perception resolutions. More ablation studies are provided in the supplementary materials.

\noindent\textbf{Optimization objectives.}
Due to the lack of abnormal training images, it's difficult to establish an explicit boundary between normal and abnormal prompt features. We adopt multiple loss functions to constrain the learning process, as shown in Tab.~\ref{tab:loss_abl}, both $\ell_{clip}$ and $\ell_{trip}$ improve the overall performance of learnable prompts. 
Then, $\ell_{mean}$ sets an explicit alignment strategy between AHPs and ALPs to improve training stability. 
Notably, $\ell_{reg}$ utilizes the CLIP feature space to constrain the learning process of abnormal prompts, maintaining the ability to learn diverse anomalies while aligning global and local semantic information.

\begin{table}\tiny
\centering
\scalebox{1.5}{
\begin{tabular}{lccccccccc}
\hline
\multirow{2}{*}{Method} & \multicolumn{3}{c}{RealIAD C1 view~\cite{wang2024realiad}} \\ \cline{2-4} 
 &  1-shot  & 2-shot & 4-shot \\ 
 \hline
 WinCLIP+~\cite{jeong2023winclip}& 82.33/94.38 & 83.21/95.11 & 83.49/95.34  \\
 PromptAD~\cite{li2024promptad}& 84.96/\textcolor{blue}{{96.21}} & 85.33/\textcolor{blue}{{96.53}} & 85.67/\textcolor{blue}{{96.73}}  \\ 
 \rowcolor{cyan!10}\textbf{FineGrainedAD}& 85.47/\textcolor{red}{96.49} & 85.83/\textcolor{red}{96.71} & 86.22/\textcolor{red}{97.16}\\ 
 \hline
\end{tabular}
} 
\caption{Quantitative comparisons of image and pixel anomaly detection in AUROC metrics on C1 view of RealIAD datasets. The best and second best pixel detection performances are in \textcolor{red}{red} and \textcolor{blue}{{blue}} colors, respectively.}
\label{tab:fs_metrics_realiad}
\end{table}

\begin{figure}[!t]
\centering
\includegraphics[width=1.0\columnwidth]{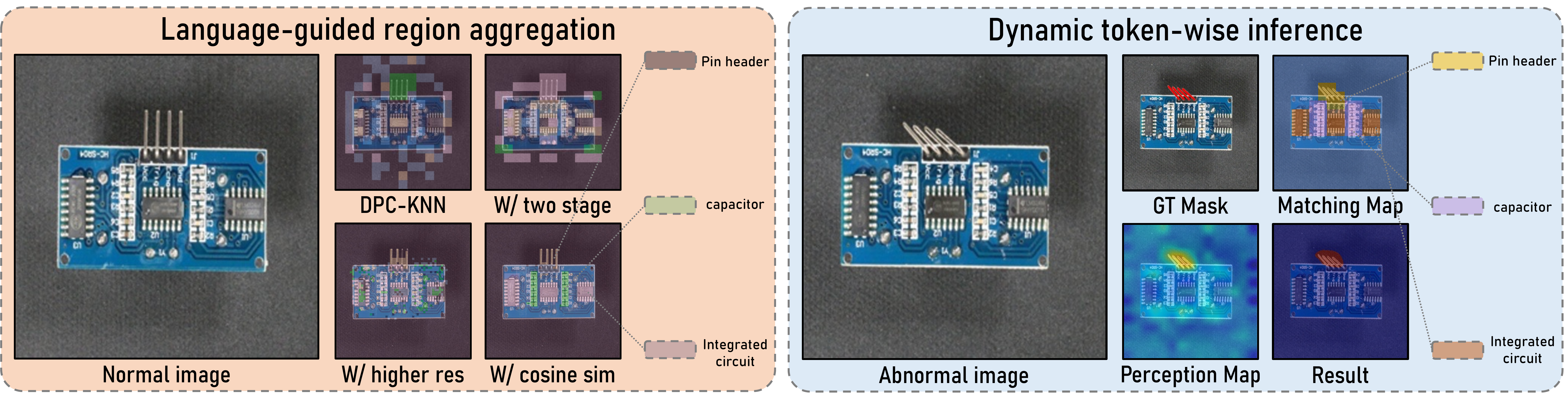}
\caption{The visualization of language-guided progressive region aggregation and dynamic token-wise inference mechanism.}
\label{fig:match_map}
\end{figure}

\noindent\textbf{Components.}
Then we investigate the effect of different components of the overall framework, as shown in Tab.~\ref{tab:comp_abl}. The multi-level prompts improve prompt granularity, but the overall performance is still limited because multi-level prompts are not aligned with the corresponding visual regions. With the alignment training, the learning ability is greatly improved. Then dynamic token-wise inference further realizes token-wise perception and fine-grained localization. Note that we take the average anomaly scores of all-level prompts within each token without dynamic token-wise inference mechanism.

\noindent\textbf{Scale up the backbone.}
To explore the reasons for the detection performance improvement more intuitively, we further replace the pre-trained backbone with other feature extractors with different perception resolution. As shown in Tab.~\ref{tab:backbone_abl}, increasing the resolution while keeping the parameters the same can bring performance improvement. 
The reason is that when the image resolution~(\ie, numbers of visual tokens) is increased, anomaly detection can be performed at a finer granularity, leading to more accurate anomaly localization. It is worth noting that the impact of the model parameters on pixel anomaly localization is relatively small compared to the performance improvement from the higher resolution, and the image-level anomaly classification ability increases steadily as the parameters increase, which can be attributed to the stronger feature extraction ability from class token in larger ViT models.

\noindent\textbf{Region aggregation and alignment effect.} To analyze the effect of the language-guided progressive region aggregation strategy more intuitively during alignment training, we visualize the aggregation map under different settings. As shown in the top of the Fig.~\ref{fig:match_map}, higher resolution patches bring significant performance improvement, while two-stage clustering achieves better separation between foreground and background. Then, replacing the clustering reference dimension improves the merging effect of similar visual component. Finally, the initial clustering selection also enhances the robustness of overall pipeline. 
Besides, to analyze the alignment effect of multi-level learnable prompts during inference, we also visualize the token-wise matching map, as shown in Fig.~\ref{fig:match_map}, different component-level prompt representations match their image regions accurately, validating the effectiveness of multi-level alignment training.

\begin{figure}[!t]
\centering
\includegraphics[width=1.0\columnwidth]{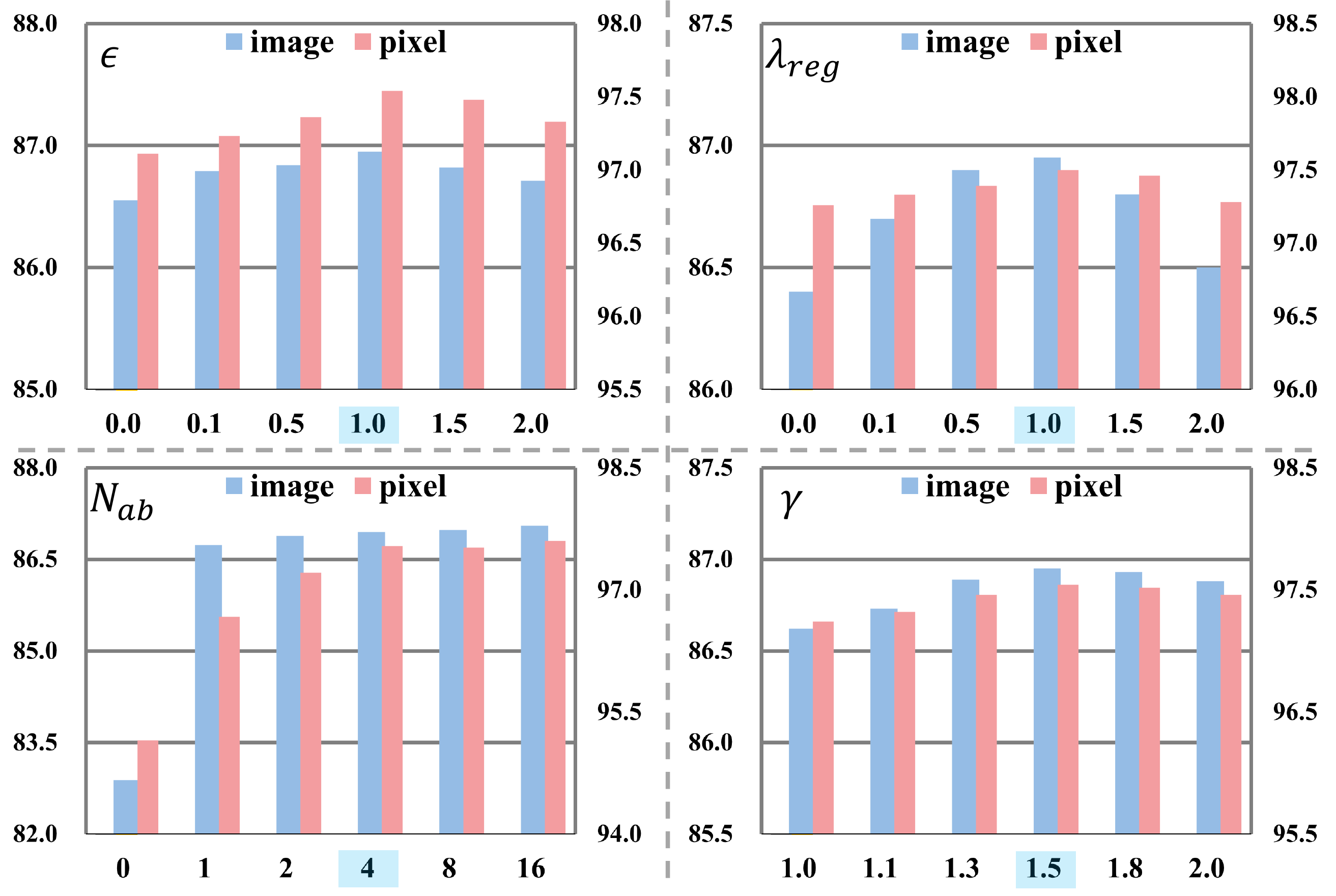}
\caption{Image/pixel AUROC on VisA dataset under 4-shot setting using different settings of $\epsilon$, $\lambda_{reg}$, $N_{ab}$, and $\gamma$.}
\label{fig:hyperP}
\end{figure}

\subsection{Hyper-parameter Analysis}
We discuss the effect of $\epsilon$ in $\ell _{trip}$ and $\lambda_{reg}$ in $\ell _{reg}$. 
As shown in Fig.~\ref{fig:hyperP}, too small $\epsilon$ cannot set explicit boundaries between normal and abnormal prompt features, while too large $\epsilon$ reduces the distribution diversity of abnormal features, both of which affect the performance. 
Meanwhile, $\lambda$ can effectively adjust the alignment strength between component-level prompt and image-level prompt, too large or too small may lead to a poor alignment effect. 
Besides, we analyze the impact of the number of abnormal learnable prompts~(ALPs) of each component $N_{ab}$. As shown in Fig.~\ref{fig:hyperP}, the model cannot learn the diverse anomaly distribution when $N$ is smaller, resulting in poor performance. When $N$ is larger than $4$, the perception capability reaches saturation. 
Besides, we also investigate the impact of learning weights $\gamma$ during alignment training, the component prompt cannot focus on the corresponding regions when the $\gamma$ is too small, and when it becomes too large, the prompt may lack the ability to perceive global semantics. 
Finally, to achieve the trade-off between performance and efficiency, we set $\epsilon$, $\lambda$, $N$, and $\gamma$ as 1, 1, 4, and 1.5 respectively. 

\section{Conclusion}
In this paper, to address the issues that the coarse perceptual granularity and semantic misalignment of existing FSAD methods, we propose Multi-Level Fine-Grained Semantic Caption~(MFSC), which construct image description architectures and utilizes the advanced MLLMs to construct fine-grained attribute descriptions based on few-shot normal images. 
Based on the MFSC, we propose a novel few-shot anomaly detection method FineGrainedAD, which consists of two components: Multi-level Learnable Prompts (MLLP) and Multi-Level Semantic Alignment~(MLSA). 
MLLP first decouples coarse-grained image-level prompts into multi-level learnable prompts to introduce fine-grained semantic information through automatic replacement and concatenation mechanisms. 
MLSA designs language-guided region aggregation and multi-level alignment training to facilitate the learnable prompts to better align with corresponding visual regions. Through dynamic token-wise inference mechanism, MLSA achieves differential token-wise anomaly detection and localization. 
Finally, FineGrainedAD achieves impressive detection performance on multiple anomaly datasets under few-shot settings. 
In future works, we will explore better fine-grained perception mechanisms and prompt learning strategies to minimize the performance gaps between zero/few-shot and full-shot AD methods.












{
    \small
    \bibliographystyle{ieeenat_fullname}
    \bibliography{{egbib}}
}

\end{document}